\documentclass{article}
\usepackage{spconf,amsmath}
\usepackage{graphicx}
\usepackage{tabularx}
\usepackage{amssymb}
\usepackage{amsfonts}
\usepackage{subfigure}
\usepackage{url}
\usepackage[
    backend=biber,
    style=ieee,
    maxbibnames=3,
    maxcitenames=3,
    doi=false,isbn=false,url=false,eprint=false
]{biblatex} 

\DeclareSourcemap{
	\maps[datatype=bibtex, overwrite=true]{
		\map{
			\step[fieldsource=booktitle,
			match=\regexp{.*Interspeech.*},
			replace={Proc. INTERSPEECH}]
			\step[fieldsource=booktitle,
			match=\regexp{.*INTERSPEECH.*},
			replace={Proc. INTERSPEECH}]
			\step[fieldsource=booktitle,
			match=\regexp{.*icassp_inpress.*},
			replace={ICASSP (in press)}]
			\step[fieldsource=booktitle,
			match=\regexp{.*International.*Conference.*on.*Acoustics,.*Speech.*and.*Signal.*Processing.*},
			replace={Proc. ICASSP}]
    		\step[fieldsource=booktitle,
 			match=\regexp{.*ICASSP Workshop.*},
 			replace={Proc. ICASSP Workshop}]    
			\step[fieldsource=booktitle,
			match=\regexp{.*International.*Conference.*on.*Learning.*Representations.*},
			replace={ICLR}]
    		\step[fieldsource=booktitle,
 			match=\regexp{.*ICLR.*},
 			replace={Proc. ICLR}]   
			\step[fieldsource=booktitle,
			match=\regexp{.*International.*Conference.*on.*Machine.*Learning.*},
			replace={ICML}]
    		\step[fieldsource=booktitle,
 			match=\regexp{.*ICML.*},
 			replace={Proc. ICML}]
			\step[fieldsource=booktitle,
			match=\regexp{.*Automatic.*Speech.*Recognition.*and.*Understanding.*},
			replace={Proc. ASRU}]
    		\step[fieldsource=booktitle,
 			match=\regexp{.*ASRU.*},
 			replace={Proc. ASRU}]
    		\step[fieldsource=booktitle,
 			match=\regexp{.*EMNLP.*},
 			replace={Proc. EMNLP}]
    		\step[fieldsource=booktitle,
 			match=\regexp{.*CVPR.*},
 			replace={Proc. CVPR}]
    		\step[fieldsource=booktitle,
 			match=\regexp{.*ECCV.*},
 			replace={Proc. ECCV}]
        	\step[fieldsource=booktitle,
 			match=\regexp{.*SIGDial.*},
 			replace={Proc. SIGDIAL}]
			\step[fieldsource=booktitle,
			match=\regexp{.*Spoken.*Language.*Technology.*},
			replace={Proc. SLT}]
    		\step[fieldsource=booktitle,
 			match=\regexp{.*SLT.*},
 			replace={Proc. SLT}]
			\step[fieldsource=booktitle,
			match=\regexp{.*Speech.*Synthesis.*Workshop.*},
			replace={Proc. SSW}]
			\step[fieldsource=booktitle,
			match=\regexp{.*workshop.*on.*speech.*synthesis.*},
			replace={Proc. SSW}]
    		\step[fieldsource=booktitle,
 			match=\regexp{.*SSW.*},
 			replace={Proc. SSW}]
    		\step[fieldsource=booktitle,
 			match=\regexp{.*ACL.*},
 			replace={Proc. ACL}]
			\step[fieldsource=booktitle,
			match=\regexp{.*Workshop.*on.* Applications.* of.* Signal.*Processing.*to.*Audio.*and.*Acoustics.*},
			replace={Proc. WASPAA}]
			\step[fieldsource=booktitle,
			match=\regexp{.*International.*Conference.*on.*Language.*Resources.*and.*Evaluation.*},
			replace={Proc. LREC}]
			\step[fieldsource=journal,
			match=\regexp{.*Spontaneous.*Speech.*Processing.*and.*Recognition},
			replace={Proc. SSPR}]
			\step[fieldsource=publisher,
			match=\regexp{.+},
			replace={{}}]
			\step[fieldsource=month,
			match=\regexp{.+},
			replace={{}}]
			\step[fieldsource=location,
			match=\regexp{.+},
			replace={{}}]
			\step[fieldsource=address,
			match=\regexp{.+},
			replace={{}}]
			\step[fieldsource=organization,
			match=\regexp{.+},
			replace={{}}]
			\step[fieldsource=doi,
			match=\regexp{.+},
			replace={{}}]
			\step[fieldsource=url,
			match=\regexp{.+},
			replace={{}}]
			\step[fieldsource=editor,
			match=\regexp{.+},
			replace={{}}]
		}
	}
}

\newcommand{\Figure}[3]{\vspace{-2mm} \includegraphics[width=#1,clip]{#2.eps} \vspace{-3mm} \caption{#3} \vspace{-4mm} \label{fig:#2}}
\newcommand{\FigureNarrow}[3]{\vspace{-2mm} \includegraphics[width=#1,clip]{#2.eps} \vspace{-6mm} \caption{#3} \vspace{-4mm} \label{fig:#2}}

\renewcommand{\Vec}[1]{\textrm{\boldmath $#1$}} 

\newcommand{\x}{ \Vec{x} } 
\newcommand{\y}{ \Vec{y} } 
  
\newcommand{\drawfigt}[4]{ 
  \begin{figure*}[#1]
  \begin{center}
  \Figure{#2}{#3}{#4}
  \end{center} 
  \end{figure*}
}
\newcommand{\drawfign}[4]{ 
  \begin{figure}[#1]
  \begin{center}
  \FigureNarrow{#2}{#3}{#4}
  \end{center} 
  \end{figure}
}

\title{StyleCap: Automatic Speaking-Style Captioning from Speech \\ Based on Speech and Language Self-supervised Learning Models}
\name{Kazuki Yamauchi$^{\star}$\sthanks{This author performed the work while at NTT Corporation as an intern.}, Yusuke Ijima$^{\dagger}$, Yuki Saito$^{\star}$}
\address{$^{\star}$The University of Tokyo, Japan, $^{\dagger}$NTT Corporation, Japan}
\begin{document}
\ninept
\setlength{\abovedisplayskip}{2pt}
\setlength{\belowdisplayskip}{2pt}
\setlength\floatsep{6pt}
\setlength\intextsep{6pt}
\setlength\textfloatsep{6pt}
\setlength{\dbltextfloatsep}{4pt}
\setlength{\dblfloatsep}{4pt}

\maketitle
\begin{abstract}
We propose \textit{StyleCap}, a method to generate natural language descriptions of speaking styles appearing in speech. Although most of conventional techniques for para-/non-linguistic information recognition focus on the category classification or the intensity estimation of pre-defined labels, they cannot provide the reasoning of the recognition result in an interpretable manner. StyleCap is a first step towards an end-to-end method for generating speaking-style prompts from speech, i.e., \textit{automatic speaking-style captioning}. StyleCap is trained with paired data of speech and natural language descriptions. We train neural networks that convert a speech representation vector into prefix vectors that are fed into a large language model (LLM)-based text decoder. We explore an appropriate text decoder and speech feature representation suitable for this new task. The experimental results demonstrate that our StyleCap leveraging richer LLMs for the text decoder, speech self-supervised learning (SSL) features, and sentence rephrasing augmentation improves the accuracy and diversity of generated speaking-style captions. Samples of speaking-style captions generated by our StyleCap are publicly available\footnote{\url{https://ntt-hilab-gensp.github.io/icassp2024stylecap/}}.
\end{abstract}

\vspace{-8pt}
\section{Introduction}
\vspace{-4pt}

Speech contains not only linguistic information but also para-/non-linguistic information~\cite{cowen2019mapping}.
The latter information, which includes the speaker's emotion and identity, can add variations to the spoken (i.e., linguistic) content and enrich human speech communication.
Hence, such information from speech needs to be automatically recognized to develop human-oriented spoken language processing technologies, such as a natural conversational agent.
Benefiting from the development of sophisticated deep learning architectures and techniques, such as Transformer~\cite{vaswani2017attention} and self-supervised learning (SSL)~\cite{shor2022trillson} for extracting meaningful feature representations from speech, the accuracy of para-/non-linguistic information recognition has been improved significantly~\cite{akccay2020speech,bai2021speaker,low2020automated}.
Moreover, the knowledge of deep learning-based technologies to recognize this information can be shared with other speech generative tasks, such as expressive text-to-speech (TTS)~\cite{guo23emodiff} and emotional voice conversion~\cite{zhou23evc}.



Another crucial factor for better para-/non-linguistic information recognition is the explainability of the recognition results.
Some applications such as healthcare will require not only the recognition result but also its reasoning~\cite{Bharati_2023}, which should be interpretable for users. However, most existing techniques focus on the category classification or the intensity estimation of labels defined by authors or dataset developers.
One possible approach to solve this issue would be representing the evidence, i.e., para-/non-linguistic information in speech, with a natural language description. Several studies have attempted to acquire the relationship between para-/non-linguistic information and natural language descriptions~\cite{dhamyal2022,pan2023gemo}. However, these methods still output class labels such as emotional states.

From a different viewpoint, generating natural language descriptions from input audio/image data can be regarded as a \textit{captioning} task.
For instance, audio captioning~\cite{mei2022automated} and image captioning~\cite{xu2023deep} are tasks to describe content information (objects and its behavior) in audio and image data, respectively.
In these tasks, the remarkable progress of large-scale pre-trained deep neural network (DNN) models~\cite{radford2021learning,elizalde2023clap,brown2020language}, has achieved the audio/image captioning performance better than conventional methods.
Unlike these tasks, this paper focuses on para-/non-linguistic information rather than content information in speech.
However, since this information cannot be directly observed in input speech~\cite{koolagudi2012emotion}, generating captions of such information can be regarded as a new challenge. 

\drawfign{t}{1.0\linewidth}{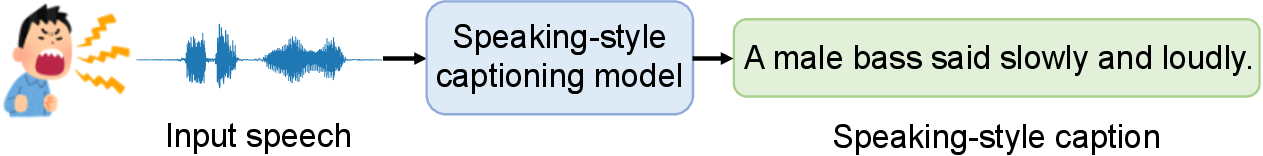}
{
Concept of automatic speaking-style captioning.
}

In this paper, we propose a deep learning technique for describing para-/non-linguistic information such as speaking style in speech with natural language, called {\it automatic speaking-style captioning} (Fig.~\ref{fig:concept}).
One way to train a DNN-based speaking-style captioning model is to use a sufficiently large dataset including many pairs of speech and sentences that describe various speaking styles, but such datasets have yet to be constructed.
Instead, we combine the existing LibriTTS and PromptSpeech corpora and build a multi-speaker speech corpus paired with the natural language instructions of speaking styles (e.g., pitch, and speed) to train an end-to-end speaking-style captioning model consisting of a speech encoder and a text decoder.
Inspired by the DNN-based method for image captioning~\cite{mokady2021clipcap}, our proposed method, \textit{StyleCap}, predicts prefix vectors fed into a large language model (LLM)-based text decoder from fixed-length speech representation vectors extracted by the speech encoder.
In this paper, we experimentally explore an appropriate text decoder and speech representation suitable for this new task.
In addition, since one speaking style can be described in various ways, automatic speaking-style captioning involves learning one-to-many mapping. 
To overcome this difficulty, we also introduce a simple data augmentation method, sentence rephrasing augmentation, that rephrases sentences using an LLM.
We conduct an automatic style-captioning experiment to assess the effectiveness of the proposed method, using evaluation metrics commonly utilized in other natural language generation tasks.
The experimental results demonstrate that our StyleCap leveraging richer LLMs for the text decoder, speech self-supervised learning (SSL) features, and sentence rephrasing augmentation improves the accuracy and diversity of generated speaking-style captions.

\drawfigt{t}{0.90\linewidth}{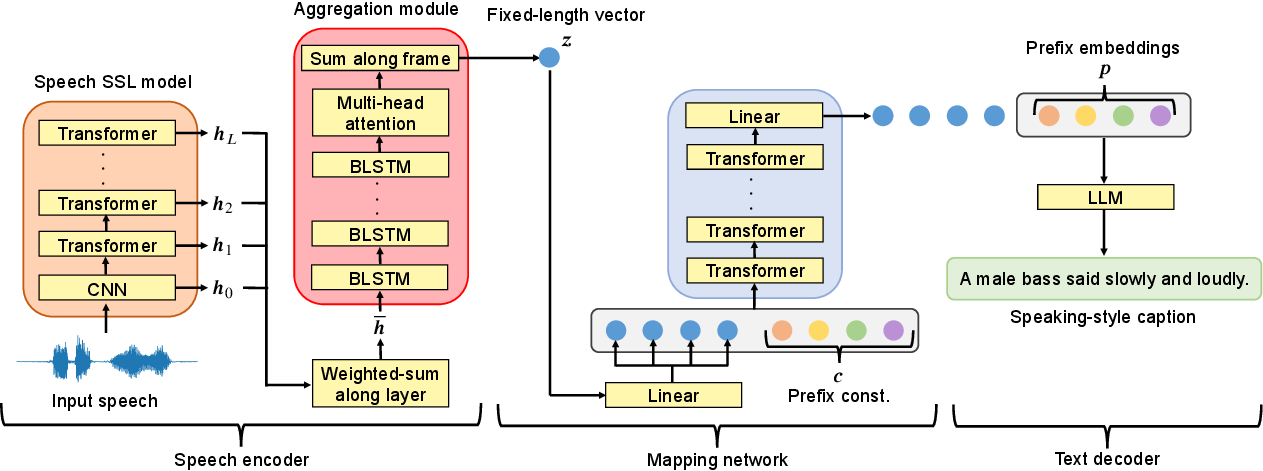}
{
DNN architecture of StyleCap, end-to-end model that automatically describes speaking style of input speech with natural language.
}

\vspace{-6pt}
\section{Methods}
\vspace{-4pt}
\subsection{Task overview}
\vspace{-4pt}

We define \textit{automatic speaking-style captioning}, a novel task that explains the speaking style of input speech with a natural language descriptions.
Specifically, a captioning model takes a single speaker's speech sample as the input and generates a sentence describing the speaker's speaking style, for instance, {\it ``The speaker's sound height is normal, but the speed is very fast, and the volume is very low.''}

Let $\x$ and $\y$ be a speech waveform and a corresponding speaking-style caption (i.e., token sequence), respectively. Because the lengths of $\x$ and $\y$ are different, one can introduce a sequence-to-sequence model widely used in spoken language processing tasks, such as Transformer~\cite{vaswani2017attention}, into this captioning task. In Section 3, we evaluate the speaking-style captioning performance of this naive Transformer-based encoder-decoder model.


\vspace{-4pt}
\subsection{StyleCap: end-to-end automatic speaking-style captioning}
\vspace{-4pt}

StyleCap incorporates two SSL models for speech and text processing and achieves the end-to-end generative modeling of speaking-style prompts.
Fig.~\ref{fig:stylecap_overview} shows an overview of StyleCap, which is inspired by ClipCap~\cite{mokady2021clipcap}, an existing model for automatic image captioning.
Specifically, StyleCap consists of three DNNs: speech encoder, text decoder, and mapping network.  


{\bf The speech encoder} extracts a fixed-length feature vector from an input speech waveform. Considering the success in various spoken language processing tasks, we incorporate a speech SSL model into the speech encoder. First, all hidden vectors of the SSL model, $\Vec{h}_{l} = [\Vec{h}_{l,1}, \ldots, \Vec{h}_{l,T}]^\top \; (l = 1, \ldots, L)$, are extracted from a speech waveform $\x$. Then, the weighted-sum of the hidden layer outputs along layers for each frame index $t$, i.e., $\Vec{\bar{h}}_t = \sum_{l}w_{l}\Vec{h}_{l,t}$, is taken, where $w_l$ is a trainable weight coefficient of the $l$th hidden layer output. Finally, the hidden vector sequence, $\Vec{\bar{h}} = [\Vec{\bar{h}}_1, \ldots, \Vec{\bar{h}}_T]^\top$, is encoded by an aggregation module consisting of a stack of bi-directional long short-term memory (BLSTM) and multi-head attention (MHA) layers, and the outputs are then taken to be summed along the frame direction to obtain a $D_{\rm z}$-dimensional fixed-length speech feature vector $\Vec{z}$.






{\bf The text decoder} leverages a pre-trained LLM and generates the speaking-style caption of the given speech waveform. Similar to the ClipCap text decoder, it first takes $K \times D_{\rm w}$ prefix embeddings $\Vec{p} = [\Vec{p}^\top_1, \ldots, \Vec{p}^\top_K]^\top$ predicted by the mapping network from the feature vector $\Vec{z}$, where $K$ and $D_{\rm w}$ denote the prefix length and the word embedding dimensionality of the LLM, respectively.
Then, the decoder autoregressively generates the tokens of speaking-style captions using the prefix embeddings as conditional vectors.

{\bf The mapping network} projects the speech encoder output $\Vec{z}$ onto the word embedding space, i.e., prefix embeddings $\Vec{p}$. It consists of a stack of Transformer layers and $K \times D_{\rm z}$ trainable prefix constant parameters $\Vec{c} = [\Vec{c}^\top_1, \ldots, \Vec{c}^\top_K]^\top$ to retrieve meaningful feature representation from $\Vec{z}$ through MHA layers.




\vspace{-4pt}
\subsection{Data augmentation by rephrasing sentences using an LLM}
\vspace{-4pt}
Because one speaking style can be described in various ways, automatic speaking-style captioning essentially requires learning a one-to-many mapping similar to TTS. To mitigate the difficulty, we introduce sentence rephrasing augmentation using an LLM to increase the diversity of speaking-style captions in the training data. Specifically, we first ask the pre-trained Llama~2-Chat (7B)\footnote{\url{https://huggingface.co/meta-llama/Llama-2-7b-chat}}~\cite{2023llama2}, which is an LLM optimized for dialogue use cases, to generate five sentences from one given sentence on the basis of the following prompt: {\it ``Rewrite the following sentence that describes someone's style of speaking in a different way, keeping the meaning of the original description. Original Description: [subject to be rephrased].''} Then, we select one sentence from the five rephrased sentences on the basis of their BERTScore~\cite{2020bertscore} values: i.e., we pick on the sentence with the highest score if the score is higher than 0.80. For instance, the description {\it ``His sound height is normal, but the speed is very fast, and the volume is very low.''} can be rephrased as {\it``Despite his normal height, his sound is incredibly fast and surprisingly quiet.''}

\begin{table*}[tb]
\centering
\caption{Automatic speaking-style captioning results. AM is the aggregation module. B@4, R, M, BS, C, and S denote BLEU@4, ROUGE-L, METEOR, BERTScore, CIDEr-D, and SPICE scores, respectively. distinct-1/-2 of reference captions are 0.020 and 0.071, respectively.}
\subtable[Results without sentence rephrasing augmentation.]{
\label{tab:result1}
  \scalebox{0.93}{
  \begin{tabular}{cccccccccc}  \hline
    Model & Speech encoder & B@4$\uparrow$ & R$\uparrow$ & M$\uparrow$ & BS$\uparrow$ & C$\uparrow$ & S$\uparrow$ & distinct-1$\uparrow$ & distinct-2$\uparrow$ \\ \hline \hline

    & Mel-spectrogram & 0.163 & 0.352 & 0.320 & 0.817 & 2.171 & 0.273 & 0.019 & 0.049 \\
    Transformer-based encoder-decoder & x-vector & 0.096 & 0.269 & 0.248 & 0.799 & 1.289 & 0.209 & 0.015 & 0.039 \\
    & WavLM & 0.253 & 0.475 & 0.456 & 0.850 & 3.239 & 0.419 & 0.022 & 0.064 \\ \hline
    & Mel-spectrogram + AM & 0.178 & 0.381 & 0.357 & 0.827 & 2.295 & 0.316 & 0.020 & 0.057 \\
    StyleCap w/ GPT-2 (proposed) & x-vector & 0.085 & 0.273 & 0.255 & 0.800 & 1.138 & 0.214 & 0.013 & 0.032 \\
    & WavLM + AM & 0.228 & 0.433 & 0.410 & 0.839 & 2.868 & 0.370 & 0.022 & 0.064 \\ \hline
    & Mel-spectrogram + AM & 0.160 & 0.358 & 0.332 & 0.821 & 2.109 & 0.295 & 0.022 & 0.066 \\
    StyleCap w/ Llama~2 (proposed) & x-vector & 0.076 & 0.262 & 0.239 & 0.799 & 1.107 & 0.213 & 0.016 & 0.042 \\
    & WavLM + AM & {\bf 0.273} & {\bf 0.497} & {\bf 0.469} & {\bf 0.855} & {\bf 3.471} & {\bf 0.434} & {\bf 0.023} & {\bf 0.073} \\ \hline
  \end{tabular}
  }
}
\subtable[Results with sentence rephrasing augmentation.]{
\label{tab:result2}
  \scalebox{0.93}{
  \begin{tabular}{cccccccccc}  \hline
    Model & Speech encoder & B@4$\uparrow$ & R$\uparrow$ & M$\uparrow$ & BS$\uparrow$ & C$\uparrow$ & S$\uparrow$ & distinct-1$\uparrow$ & distinct-2$\uparrow$ \\ \hline \hline
    & Mel-spectrogram & 0.140 & 0.332 & 0.303 & 0.814 & 1.847 & 0.239 & 0.018 & 0.046 \\
    Transformer-based encoder-decoder & x-vector & 0.071 & 0.244 & 0.212 & 0.792 & 1.046 & 0.191 & 0.012 & 0.027 \\
    & WavLM & 0.246 & 0.464 & 0.441 & 0.848 & 3.172 & 0.404 & 0.021 & 0.059 \\ \hline
    & Mel-spectrogram + AM & 0.164 & 0.368 & 0.334 & 0.822 & 2.122 & 0.294 & 0.021 & 0.063 \\
    StyleCap w/ GPT-2 (proposed) & x-vector & 0.068 & 0.260 & 0.237 & 0.798 & 0.895 & 0.210 & 0.013 & 0.033 \\
    & WavLM + AM & 0.239 & 0.470 & 0.439 & 0.848 & 3.056 & 0.403 & 0.022 & 0.068 \\ \hline
    & Mel-spectrogram + AM & 0.165 & 0.353 & 0.327 & 0.818 & 2.131 & 0.279 & 0.024 & 0.065 \\
    StyleCap w/ Llama~2 (proposed) & x-vector & 0.084 & 0.259 & 0.237 & 0.796 & 1.157 & 0.212 & 0.014 & 0.034 \\
    & WavLM + AM & {\bf 0.279} & {\bf 0.507} & {\bf 0.479} & {\bf 0.857} & {\bf 3.594} & {\bf 0.447} & {\bf 0.027} & {\bf 0.079} \\ \hline
  \end{tabular}
  }
}
\end{table*}

\vspace{-6pt}
\section{Experiments}
\vspace{-4pt}
\subsection{Dataset}
\vspace{-4pt}

We used PromptSpeech\footnote{\url{https://speechresearch.github.io/prompttts}}, which includes natural language instructions (style prompt) of various speaking styles.
This dataset was constructed for PromptTTS~\cite{guo2022ptompttts} that can synthesize speech in accordance with a style prompt.
We used the training subset of PromptSpeech real version, which includes 26,588 human-annotated style prompts of multi-speakers' speech samples from LibriTTS~\cite{zen2019libri}.
We divided the 26,588 prompts into training (24,953 by 1,113 speakers), development (857 by 40), and test (778 by 38) subsets and paired them with the corresponding speech samples from LibriTTS.
Although PromptSpeech also includes categorical style factors that indicate class labels regarding gender, pitch, speed, and volume, we did not use them to train the captioning models.


\vspace{-4pt}
\subsection{Experimental conditions}
\vspace{-4pt}
As for the speech encoder, we used three types of speech feature representations: mel-spectrogram, speaker embeddings for speaker verification, and hidden vectors of a speech SSL model for the performance comparison. We extracted 80-dimensional mel-spectrograms from speech waveforms with 10~ms frame shift. We also used WavLM {\sc BASE+}\footnote{\url{https://huggingface.co/microsoft/wavlm-base-plus}}~\cite{Chen2021WavLM} as the speech SSL model and extracted weighted-sum representation for each frame. To obtain a fixed-length vector representation for the mel-spectrogram and speech SSL model, a stack of 4-layer BLSTMs and MHA with 8 heads was used as an aggregation module.
As for the speaker embeddings, 512-dimensional x-vectors~\cite{snyder2018xvector} obtained from the pre-trained WavLM {\sc BASE+} for speaker verification\footnote{\url{https://huggingface.co/microsoft/wavlm-base-plus-sv}} was used.
As for the text decoder, we used two types of pre-trained LLMs: GPT-2\footnote{\url{https://huggingface.co/gpt2}}~\cite{radford2019language} and Llama~2 (7B)\footnote{\url{https://huggingface.co/meta-llama/Llama-2-7b}}~\cite{2023llama2}.
The former is the same setting as ClipCap~\cite{mokady2021clipcap}, while the latter can be regarded as the richer one. 
The numbers of model parameters for GPT-2 and Llama~2 were 125M and 7B, respectively. The dimensions of word embeddings for each model were 768 and 4,096, respectively.
The mapping network consisted of 8-layer Transformer-encoders. The prefix length and dropout ratio were set to 40 and 0.2, respectively. These parameters were empirically determined.
During the model training, we only trained the mapping network, the aggregation module in the speech encoder, and the weight coefficient for each layer in WavLM (i.e., $w_l$) used in the SSL feature aggregation process. The model parameters of LLMs and WavLM were frozen.
All modules are trained in an end-to-end manner using the cross entropy loss between the generated captions and the ground-truth captions.
Each model was trained without sentence rephrasing augmentation at 20 epochs or with the augmentation at 10 epochs because the augmentation doubles the data size. The batch size was set to 16.

We also trained a naive Transformer-based encoder-decoder model as the baseline system. The input features were extracted by the almost same speech encoder as those explained in the previous paragraph, but the aggregation module was replaced by an attention layer to align the lengths of speech features and tokens in a generated caption.
The encoder had 12 encoder blocks. The decoder had a token embedding layer followed by 6 decoder blocks. We adopted MHA with 4 heads of 256 dimensions for both encoder and decoder. The implementation was based on Huggingface Transformers\footnote{\url{https://huggingface.co/docs/transformers/model_doc/speech_to_text}}.


As evaluation metrics for captioning accuracy, we used BLEU~\cite{2002bleu}, ROUGE-L~\cite{lin2004rouge}, METEOR~\cite{Banerjee2005meteor}, BERTScore~\cite{2020bertscore}, CIDEr-D~\cite{Vedantam2015cider}, and SPICE~\cite{Anderson2016spice}, between generated and annotated captions referring to other natural language generation tasks.
We also used distinct-1/-2~\cite{2016distinct} to evaluate the diversity of generated captions.

\vspace{-4pt}
\subsection{Experimental results}
\vspace{-4pt}

Table~\ref{tab:result1} shows experimental results without sentence rephrasing augmentation. First, a performance comparison by the speech encoder difference showed WavLM performed the best in this as well as other speech tasks. On the other hand, x-vectors did not work well because they were mainly trained to represent speaker characteristics rather than speaking styles such as speech rhythm~\cite{fujita2023zeroshot}.
In addition, the use of Llama~2 for the text decoder performed better than GPT-2 when employing WavLM for the speech encoder, which demonstrates that leveraging of richer LLM-based text decoders is a crucial factor to improve the captioning performance of StyleCap.
On the other hand, Llama~2 did not necessarily improve the performance when mel-spectrogram was used as the input feature. One reason for this would be overfitting to the training data. The word embedding dimension of Llama~2 was 4,096-dimensional embeddings, which greatly outnumbers that of GPT-2~(768). Therefore, the speech encoder with the mel-spectrogram input might suffer from the high dimensionality of Llama~2 word embeddings and result in overfitting.
In contrast, WavLM prevented such overfitting thanks to pre-training using a massive amount of speech data. 
In summary, StyleCap employing WavLM and Llama~2 outperformed naive Transformer-based encoder-decoder models.

Table~\ref{tab:result2} shows experimental results with sentence rephrasing augmentation.
Overall tendencies were similar to the case without the augmentation.
We can see that sentence rephrasing augmentation can improve the captioning performance of StyleCap employing WavLM. In other words, sentence rephrasing augmentation can make StyleCap to generate more diverse and accurate speaking-style captions, as shown in the improved distinct-1/-2 and other captioning results. These results indicate that sentence rephrasing augmentation is effective to deal with the difficulty of speaking-style captioning, i.e., learning one-to-many mapping.
In contrast, the performance of Transformer-based encoder-decoder models could not be improved.

\begin{table}[tb]
\centering
\caption{The accuracy (\%) of style factor classifications with embedding for each speech encoder. P, S, and V indicate Pitch, Speed, and Volume respectively.}
\label{tab:result3}
  \scalebox{0.95}{
  \begin{tabular}{cccccc}  \hline
    Speech encoder  & Gender & P & S & V & Avg. \\ \hline \hline
    Mel-spectrogram + AM & 93.8 & 62.1 & 67.8 & 54.7 & 69.6 \\
    x-vector & 94.0 & 40.5 & 44.0 & 49.4 & 57.0 \\
    WavLM + AM & 91.0 & 61.0 & 85.2 & 69.9 & 76.8 \\ \hline
  \end{tabular}
}
\end{table}

\begin{table}[tb]
\centering
\caption{Evaluation scores with changing prefix length.}
\label{tab:result4}
  \scalebox{0.95}{
  \begin{tabular}{c||cccccc}  \hline
    Prefix length & 1 & 2 & 5 & 10 & 40 & 60 \\ \hline
    METEOR & 0.419 & 0.437 & 0.468 & 0.464 & {\bf 0.479} & 0.459 \\
    BERTScore & 0.839 & 0.845 & 0.853 & 0.853 & {\bf 0.857} & 0.852 \\ \hline
  \end{tabular}
}
\end{table}


\vspace{-4pt}
\subsection{Discussion}
\vspace{-4pt}

\subsubsection{Analysis of StyleCap behavior}
\vspace{-4pt}
To further understand the methods, we performed analysis from two aspects: the property of fixed-length vectors extracted by the speech encoders and the captioning performance from the vectors.

We first performed style factor classification using the learned fixed-length vectors. As described in Sect. 3.1, PromptSpeech also includes class labels indicating four style factors: gender, pitch, speed, and volume. The numbers of class labels for gender and the others are two (male/female), and three (low/mid/high), respectively.
We trained a simple linear layer to classify the four style factors from a fixed-length vector extracted by the speech encoder of each trained StyleCap.
In this analysis, trained StyleCap employing Llama~2 for the text decoder and sentence rephrasing augmentation for the training were used.
Table~\ref{tab:result3} lists the classification accuracy. 
We can see that all speech encoders can classify each style factor to some extent. This implies that StyleCap can acquire speaking-style-related information from speech with natural language descriptions only. In other words, such information can be obtained without class labels indicating style factors.
Comparing the speech encoders, WavLM achieved the highest classification performance, while x-vector could not capture para-linguistic information except for gender. Similar tendency was also reported in the previous work~\cite{fujita2023zeroshot}.

\drawfign{t}{0.8\linewidth}{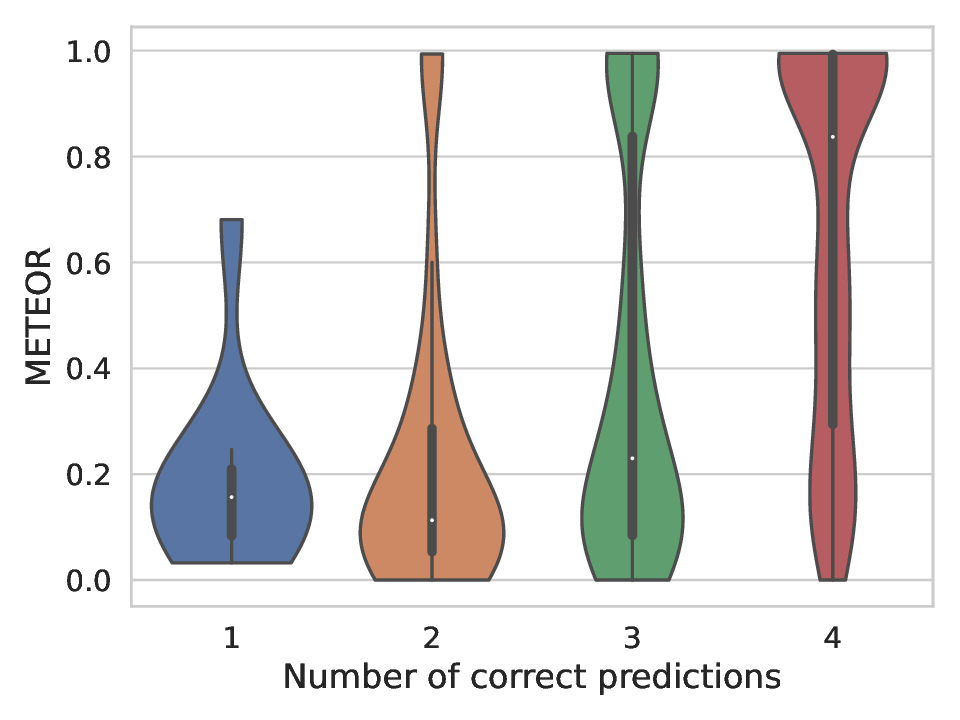}
{
Violin plot of METEOR by the performance of style factor classifications.
}

We next analyzed the relationship between the captioning performance and the fixed-length vector extracted by the speech encoder employing WavLM.
Fig.~\ref{fig:dist_meteor} shows the violin plot of METEOR aggregated by the style factor classification performance (i.e., the number of correct predictions from 1 to 4)\footnote{We did not observe the complete misclassification whose number of correct predictions was 0 in the experiments.} described in the previous paragraph.
From the results, we can see that the captioning performance strongly depends on the style factor classification performance. Especially, the misclassification of even one factor can considerably degrade the captioning performance.
Similar tendencies were also found in terms of other metrics such as BERTScore.
These results indicate that capturing adequate para-/non-linguistic information by the speech encoder is an essential factor to further improve performance.
Although WavLM and the simple aggregation module were used in this paper, another possible approach is to use CLAP~\cite{pan2023gemo,elizalde2023clap} trained from various para-/non-linguistic information.

\vspace{-4pt}
\subsubsection{Ablation study for mapping network}
\vspace{-4pt}

We finally performed an ablation study for the mapping network.
To investigate the performance by changing the prefix length ($K$) of the mapping network, we set the prefix length to 1, 2, 5, 10, 40, and 60, respectively.
In this experiment, StyleCap employing WavLM, Llama~2, and sentence rephrasing augmentation was used.
Table~\ref{tab:result4} shows METEOR and BERTScore only by changing the prefix length due to the space limitation.
As we can see, the performance is getting better with the longer prefix length.
The shorter prefix lengths, especially 1 and 2, tends to worsen the performances because the number of trainable model parameters are limited.
In addition, the too longer prefix also degrades the performance due to overfitting to training data.
These tendencies were similar to ClipCap~\cite{mokady2021clipcap}.

\vspace{-6pt}
\section{Conclusions}
\vspace{-4pt}

In this paper, we proposed StyleCap, a method to generate natural language descriptions of speaking styles appearing in speech, which we call the automatic speaking-style captioning task. As a first step to this end, we explored an appropriate large language model (LLM)-based text decoder and speech feature representation suitable for this task.
The experimental results demonstrated that our StyleCap leveraging richer LLMs for the text decoder, speech self-supervised learning (SSL) features, and sentence rephrasing augmentation improved the accuracy and diversity of generated speaking-style captions.
Although this paper focused on speaking styles appearing in speech, we believe that the proposed approach will be easily applicable for other para-/non-linguistic information such as emotional states and mental illnesses.
Applying it to such information, which includes constructing pairs of speech and natural language descriptions, is for future work. 
We will also explore more suitable evaluation metrics for this task to growth the research area regarding the speaking style captioning.
The evaluation metric using a larger LLM~\cite{liu2023gpteval} including the prompt engineering for each task would be a possible approach.

\textbf{Acknowledgements:}
This work was supported by JST, ACT-X Grant Number JPMJAX23CB, Japan.

\newpage

\printbibliography

@article{cowen2019mapping,
  title={Mapping 24 emotions conveyed by brief human vocalization},
  author={Cowen, Alan S and Elfenbein, Hillary Anger and Laukka, Petri and Keltner, Dacher},
  journal={American Psychologist},
  volume={74},
  number={6},
  pages={698--712},
  year={2019},
  publisher={American Psychological Association}
}

@inproceedings{vaswani2017attention,
  title={Attention is all you need},
  author={Vaswani, Ashish and Shazeer, Noam and Parmar, Niki and Uszkoreit, Jakob and Jones, Llion and Gomez, Aidan N and Kaiser, {\L}ukasz and Polosukhin, Illia},
  booktitle={Proc. NIPS},
  year={2017}
}

@INPROCEEDINGS{fujita2023zeroshot,
  author={Fujita, Kenichi and Ashihara, Takanori and Kanagawa, Hiroki and Moriya, Takafumi and Ijima, Yusuke},
  booktitle={Proc. ICASSP SASB Workshop}, 
  title={Zero-Shot Text-to-Speech Synthesis Conditioned Using Self-Supervised Speech Representation Model},
  year={2023},
  doi={10.1109/ICASSPW59220.2023.10193459}
}

@article{liu2023gpteval,
  title={{G-E}val: {NLG} evaluation using {GPT}-4 with better human alignment},
  author={Liu, Yang and Iter, Dan and Xu, Yichong and Wang, Shuohang and Xu, Ruochen and Zhu, Chenguang},
  journal={arXiv preprint arXiv:2303.16634},
  year={2023}
}

@inproceedings{radford2021learning,
  title={Learning transferable visual models from natural language supervision},
  author={Radford, Alec and Kim, Jong Wook and Hallacy, Chris and Ramesh, Aditya and Goh, Gabriel and Agarwal, Sandhini and Sastry, Girish and Askell, Amanda and Mishkin, Pamela and Clark, Jack and others},
  booktitle={Proc. ICML},
  pages={8748--8763},
  year={2021},
  organization={PMLR}
}

@inproceedings{elizalde2023clap,
  title={{CLAP}: Learning audio concepts from natural language supervision},
  author={Benjamin, Elizalde and Soham, Deshmukh and M. A. Ismail and Huaming, Wang},
  booktitle={Proc. ICASSP},
  year={2023},
  organization={IEEE}
}

@inproceedings{shor2022trillson,
  title={Universal Paralinguistic Speech Representations Using self-Supervised Conformers},
  author={J. Shor and A. Jansen and W. Han and D. Park and Y. Zhang},
  booktitle={Proc. ICASSP},
  year={2022},
  pages={3169--3173},
  organization={IEEE}
}

@INPROCEEDINGS{guo23emodiff,
  author={Guo, Yiwei and Du, Chenpeng and Chen, Xie and Yu, Kai},
  booktitle={Proc. ICASSP}, 
  title={{EmoDiff}: Intensity Controllable Emotional Text-to-Speech with Soft-Label Guidance}, 
  year={2023},
  volume={},
  number={},
  doi={10.1109/ICASSP49357.2023.10095621}
}

@ARTICLE{zhou23evc,
  author={Zhou, Kun and Sisman, Berrak and Rana, Rajib and Schuller, Björn W. and Li, Haizhou},
  journal={IEEE Transactions on Affective Computing}, 
  title={Emotion Intensity and its Control for Emotional Voice Conversion}, 
  year={2023},
  volume={14},
  number={1},
  pages={31--48},
  doi={10.1109/TAFFC.2022.3175578}}

@article{koolagudi2012emotion,
  title={Emotion recognition from speech: a review},
  author={Koolagudi, Shashidhar G and Rao, K Sreenivasa},
  journal={International journal of speech technology},
  volume={15},
  pages={99--117},
  year={2012},
  publisher={Springer}
}

@inproceedings{brown2020language,
  title={Language models are few-shot learners},
  author={Brown, Tom and Mann, Benjamin and Ryder, Nick and Subbiah, Melanie and Kaplan, Jared D and Dhariwal, Prafulla and Neelakantan, Arvind and Shyam, Pranav and Sastry, Girish and Askell, Amanda and others},
  booktitle={Proc. NeurIPS},
  year={2020}
}

@article{mei2022automated,
  title={Automated audio captioning: an overview of recent progress and new challenges},
  author={Mei, Xinhao and Liu, Xubo and Plumbley, Mark D and Wang, Wenwu},
  journal={EURASIP Journal on Audio, Speech, and Music Processing},
  volume={2022},
  number={1},
  pages={1--18},
  year={2022},
  publisher={SpringerOpen}
}

@article{xu2023deep,
  title={Deep Image Captioning: A Review of Methods, Trends and Future Challenges},
  author={Xu, Liming and Tang, Quan and Lv, Jiancheng and Zheng, Bochuan and Zeng, Xianhua and Li, Weisheng},
  journal={Neurocomputing},
  pages={126287},
  volume={546},
  year={2023},
  publisher={Elsevier}
}

@article{bai2021speaker,
  title={Speaker recognition based on deep learning: An overview},
  author={Bai, Zhongxin and Zhang, Xiao-Lei},
  journal={Neural Networks},
  volume={140},
  pages={65--99},
  year={2021},
  publisher={Elsevier}
}

@article{akccay2020speech,
  title={Speech emotion recognition: Emotional models, databases, features, preprocessing methods, supporting modalities, and classifiers},
  author={M. B. Ak{\c{c}}ay and K. O{\u{g}}uz},
  journal={Speech Communication},
  volume={116},
  pages={56--76},
  year={2020},
  publisher={Elsevier}
}

@article{low2020automated,
  title={Automated assessment of psychiatric disorders using speech: A systematic review},
  author={D. M. Low and K. H. Bentley and S. S. Ghosh},
  journal={Laryngoscope Investigative Otolaryngology},
  volume={5},
  number={1},
  pages={96--116},
  year={2020},
  publisher={Wiley Online Library}
}

@article{pan2023gemo,
  title={{GEmo-CLAP}: Gender-Attribute-Enhanced Contrastive Language-Audio Pretraining for Speech Emotion Recognition},
  author={Pan, Yu and Hu, Yanni and Yang, Yuguang and Yao, Jixun and Fei, Wen and Ma, Lei and Lu, Heng},
  journal={arXiv preprint arXiv: 2306.07848},
  year={2023}
}

@article{Bharati_2023,
	doi = {10.1109/tai.2023.3266418},
	url = {https://doi.org/10.1109%2Ftai.2023.3266418},
	year = 2023,
	publisher = {Institute of Electrical and Electronics Engineers ({IEEE})},
	pages = {1--15},
	author = {Subrato Bharati and M. Rubaiyat Hossain Mondal and Prajoy Podder},
	title = {A Review on Explainable Artificial Intelligence for Healthcare: Why, How, and When?},
	journal = {{IEEE} Transactions on Artificial Intelligence}
}

@inproceedings{guo2022ptompttts,
  title={{PromptTTS}: Controllable Text-to-Speech with Text Descriptions},
  author={Zhifang Guo and Yichong Leng and Yihan Wu and Sheng Zhao and Xu Tan},
  booktitle={Proc. ICASSP},
  year={2023}
}

@inproceedings{zen2019libri,
 author = {Heiga Zen and Viet Dang and Rob Clark and Yu Zhang and Ron J. Weiss and {et~al.}},
 booktitle = {INTERSPEECH},
 title = {{LibriTTS}: A corpus derived from {LibriSpeech} for text-to-speech},
 pages = {1526--1530},
 year = {2019}
}

@inproceedings{lin2004rouge,
 author = {C.-Y. Lin},
 booktitle = {Proc. Workshop on Text Summarization Branches Out},
 title = {{ROUGE}: A package for automatic evaluation of summaries},
 pages = {74--81},
 year = {2004}
}

@inproceedings{Banerjee2005meteor,
 author = {S. Banerjee and A. Lavie},
 booktitle = {Proc. ACL Workshop on Intrinsic and Extrinsic Evaluation Measures for Nachine Translation and/or Summarization},
 title = {{METEOR}: An Automatic Metric for {MT} Evaluation with Improved Correlation with Human Judgments},
 pages = {65--72},
 year = {2005}
}

@inproceedings{Vedantam2015cider,
 author = {R. Vedantam and C. Lawrence Zitnick and D. Parikh},
 booktitle = {CVPR},
 title = {{CIDEr}: Consensus-based image description evaluation},
 pages = {4566--4575},
 year = {2015}
}

@inproceedings{Anderson2016spice,
 author = {P. Anderson and B. Fernando and M. Johnson and S. Gould},
 booktitle = {ECCV},
 title = {{SPICE}: Semantic propositional image caption evaluation},
 pages = {382--398},
 year = {2016}
}

@inproceedings{2020bertscore,
  title={{BERTScore}: Evaluating Text Generation with {BERT}},
  author={Tianyi Zhang and Varsha Kishore and Felix Wu and Kilian Q. Weinberger and Yoav Artzi},
  booktitle={ICLR},
  year={2020},
  url={https://openreview.net/forum?id=SkeHuCVFDr}
}

@inproceedings{2016distinct,
 author = {Jiwei Li and Michel Galley and Chris Brockett and Jianfeng Gao and Bill Dolan},
 booktitle = {NAACL-HLT},
 title = {A Diversity Promoting Objective Function for Neural Conversation Models},
 pages = {110--119},
 year = {2016}
}

@inproceedings{2002bleu,
 author = {Kishore Papineni and Salim Roukos and Todd Ward and Wei-Jing Zhu},
 booktitle = {ACL},
 title = {{BLEU}: A method for automatic evaluation of machine translation},
 pages = {311--318},
 year = {2002}
}

@article{mokady2021clipcap,
  title={{ClipCap}: {CLIP} Prefix for Image Captioning},
  author={Ron Mokady and Amir Hertz and Amit H. Bermano},
  journal={arXiv preprint arXiv: 2111.09734},
  year={2021}
}

@article{Chen2021WavLM,
  title = {{WavLM}: Large-Scale Self-Supervised Pre-training for Full Stack Speech Processing},
  author  = {Sanyuan Chen and Chengyi Wang and Zhengyang Chen and Yu Wu and Shujie Liu and Zhuo Chen and Jinyu Li and Naoyuki Kanda and Takuya Yoshioka and Xiong Xiao and Jian Wu and Long Zhou and Shuo Ren and Yanmin Qian and Yao Qian and Jian Wu and Michael Zeng and Furu Wei},
  journal={IEEE Journal of Selected Topics in Signal Processing},
  volume={16},
  number={6},
  pages={1505--1518},
  year={2022}
}

@inproceedings{snyder2018xvector,
  author={D. Snyder and D. Garcia-Romero and G. Sell and D. Povey and S. Khudanpur},
  title={{X-Vectors}: Robust DNN Embeddings for Speaker Recognition},
  year=2018,
  booktitle={Proc. ICASSP},
  pages={5329--5333},
  doi={10.1109/ICASSP.2018.8461375}
}

@article{radford2019language,
  title={Language Models are Unsupervised Multitask Learners},
  author={Radford, Alec and Wu, Jeff and Child, Rewon and Luan, David and Amodei, Dario and Sutskever, Ilya},
  year={2019}
}

@article{2023llama2,
  title={{Llama 2}: Open Foundation and Fine-Tuned Chat Models},
  author={Hugo Touvron and Louis Martin and Kevin Stone and Peter Albert and Amjad Almahairi and Yasmine Babaei and Nikolay Bashlykov and Soumya Batra and Prajjwal Bhargava and Shruti Bhosale and Dan Bikel and Lukas Blecher and Cristian Canton Ferrer and Moya Chen and Guillem Cucurull and David Esiobu and Jude Fernandes and Jeremy Fu and Wenyin Fu and Brian Fuller and Cynthia Gao and Vedanuj Goswami and Naman Goyal and Anthony Hartshorn and Saghar Hosseini and Rui Hou and Hakan Inan and Marcin Kardas and Viktor Kerkez and Madian Khabsa and Isabel Kloumann and Artem Korenev and Punit Singh Koura and Marie-Anne Lachaux and Thibaut Lavril and Jenya Lee and Diana Liskovich and Yinghai Lu and Yuning Mao and Xavier Martinet and Todor Mihaylov and Pushkar Mishra, Igor Molybog and Yixin Nie and Andrew Poulton and Jeremy Reizenstein and Rashi Rungta and Kalyan Saladi and Alan Schelten and Ruan Silva and Eric Michael Smith and Ranjan Subramanian and Xiaoqing Ellen Tan and Binh Tang and Ross Taylor and Adina Williams and Jian Xiang Kuan and Puxin Xu and Zheng Yan and Iliyan Zarov and Yuchen Zhang and Angela Fan and Melanie Kambadur and Sharan Narang and Aurelien Rodriguez and Robert Stojnic and Sergey Edunov and Thomas Scialom},
  journal={arXiv preprint arXiv:2307.09288},
  year={2023}
}

@article{dhamyal2022,
  title={Describing emotions with acoustic property prompts for speech emotion recognition},
  author={Hira Dhamyal and Benjamin Elizalde and Soham Deshmukh and Huaming Wang and Bhiksha Raj and Rita Singh},
  journal={arXiv preprint arXiv:2211.07737},
  year={2022}
}

\end{document}